# Efficient Multi-View Fusion and Flexible Adaptation to View Missing in Cardiovascular System Signals

Qihan Hu, Daomiao Wang, Hong Wu, Jian Liu, Cuiwei Yang

*Abstract*—**The progression of deep learning and the widespread adoption of sensors have facilitated automatic multi-view fusion (MVF) about the cardiovascular system (CVS) signals. However, prevalent MVF model architecture often amalgamates CVS signals from the same temporal step but different views into a unified representation, disregarding the asynchronous nature of cardiovascular events and the inherent heterogeneity across views, leading to catastrophic view confusion. Efficient training strategies specifically tailored for MVF models to attain comprehensive representations need simultaneous consideration. Crucially, real-world data frequently arrives with incomplete views, an aspect rarely noticed by researchers. Thus, the View-Centric Transformer (VCT) and Multitask Masked Autoencoder (M2AE) are specifically designed to emphasize the centrality of each view and harness unlabeled data to achieve superior fused representations. Additionally, we systematically define the missing-view problem for the first time and introduce prompt techniques to aid pretrained MVF models in flexibly adapting to various missing-view scenarios. Rigorous experiments involving atrial fibrillation detection, blood pressure estimation, and sleep staging—typical health monitoring tasks—demonstrate the remarkable advantage of our method in MVF compared to prevailing methodologies. Notably, the prompt technique requires finetuning less than 3% of the entire model's data, substantially fortifying the model's resilience to view missing while circumventing the need for complete retraining. The results demonstrate the effectiveness of our approaches, highlighting their potential for practical applications in cardiovascular health monitoring. Codes and models are released at URL.**

## 1. INTRODUCTION

CARDIOVASCULAR system (CVS) is usually observed through various sensors, generating multi-view CVS data that offer an effective strategy for investigating this intricate system. Medical imaging, for instance, provides detailed geometric insights into the CVS's structure. The classical 12-lead electrocardiogram (ECG) captures the cardiac behavior across 12 spatial orientations, portraying the perspective of cardiac electrical activity. Besides, photoplethysmography (PPG) offers insights into blood flow status and peripheral vascular characteristics from a hemodynamic standpoint. Thus, fusing these CVS data from different views can facilitate a deeper understanding of CVS characteristics, unlocking substantial potential for diverse healthcare applications by extracting potential patterns and interactions within the data [1][2]. Single-lead ECG and PPG offer a comprehensive understanding of CVS from complementary perspectives and are widely integrated into various wearable devices due to their portability and low cost. Therefore, we focus on these two prevalent views within the CVS.

Despite significant efforts devoted to multi-view fusion (MVF) for achieving remarkable performance in specific health monitoring tasks [3][4][5], these endeavors often assume complete data availability from each view. However, real-world scenarios often deviate from this assumption due to diverse factors, such as the acquisition environment and device characteristics, resulting in the so-called missing-view problem. For instance, while a daily wristwatch actively captures PPG signals, obtaining ECG signals typically requires the wearer to place their hand on the crown; a passive measurement method often leads to missing ECG data. Conversely, hospitals' preferences for ECG measurements can lead to the absence of PPG data. The performance of a well-trained MVF model may degrade significantly when the data is view-incomplete.

Thus, two urgent questions naturally emerge: *Is the current MVF method optimal for effectively fusing multi-view CVS information*? *How can a well-established MVF model adapt flexibly to complex missing-view scenarios*?

End-to-end deep learning (DL) has garnered significant attention within physiological signal processing due to its adaptability across diverse datasets and reduced reliance on prior knowledge. Within this approach, the model architecture defines the methodology for extracting valuable representations from the data, while the training strategy dictates the quality and

This work was supported of by National Nature Science Foundation of China under Grant 62371138. (*Corresponding author: Cuiwei Yang).
Qihan Hu, Daomiao Wang, Hong Wu, Liu Jian and Cuiwei Yang are with the Center for Biomedical Engineering, School of Information Science and Technology, Fudan University, Shanghai, 200433, P.R. China (e-mail: qhhu21@m.fudan.edu.cn;wangdm23@m.fudan.edu.cn;hongwu22@m.fudan.edu.cn; liuj22@m.fudan.edu.cn).
Cuiwei Yang is also with the Key Laboratory of Medical Imaging Computing and Computer Assisted Intervention of Shanghai, 200093, P.R. China (e-mail: yangcw@fudan.edu.cn).

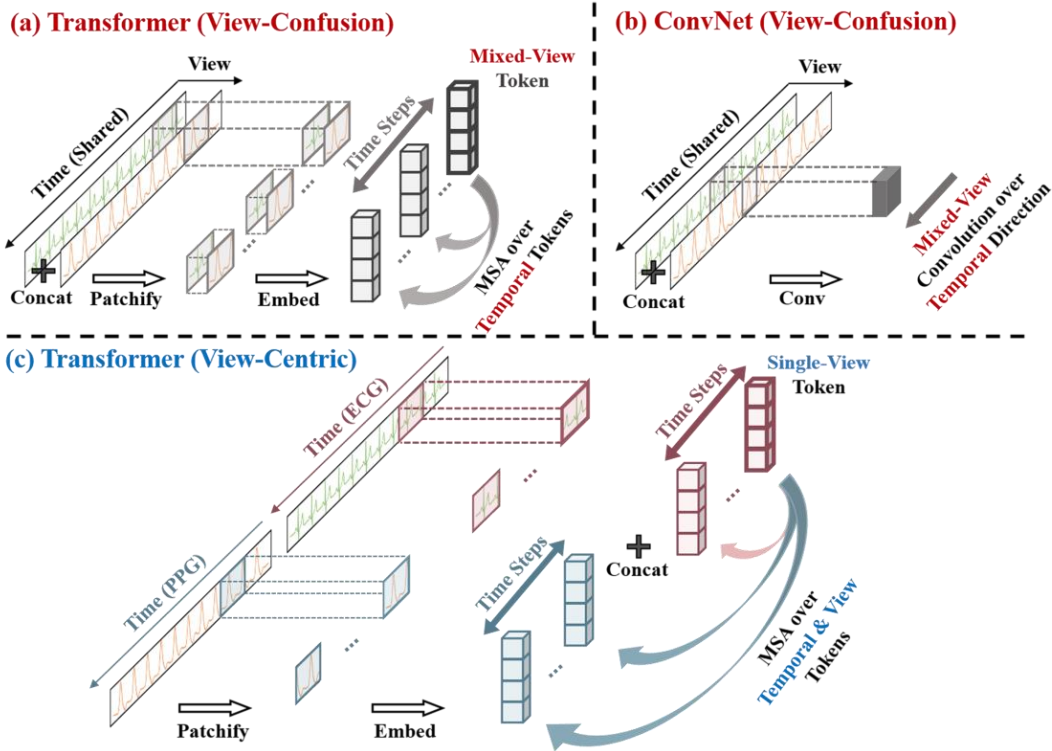

Fig. 1. Comparison between the common architecture (top) and the proposed VCT (bottom). Unlike common methods, which extract mixed-view representation from the same step, VCT embeds the sub-series of each view to the single-view token, such that multi-view and temporal interactions can be captured by the attention mechanism.

optimization direction of these representations. These two pivotal components significantly influence the effectiveness of the MVF representation obtained by DL methods.

We note several limitations in existing MVF architecture in the CVS context. First, the common practice involves extracting temporal interactions from each sample point, considered as the basic element, by popular Transformer architecture [6][7][8], which leads to high computational complexity due to the multi-head self-attention (MSA) [9]. Additionally, given that CVS signals frequently exhibit periodic waveform patterns, utilizing sub-series granularity for modeling appears more plausible than relying on sample-point granularity. Another limitation lies in the common practice of concatenating multiple signals into a multi-channel time series [10][11]. Due to the intricate physical mechanisms within the CVS, multi-view CVS signals always display timing misalignment concerning cardiovascular events. For instance, the peak of the PPG waveform always lags behind the R wave of the ECG waveform within the same heartbeat [12]. Consequently, Fig. 1(a) illustrates that forcibly embedding these sub-series, signifying distinct cardiovascular information, into a unified mixed-view token disrupts rational multi-view correlations and inadvertently deteriorates cross-view interactions. Notably, the widely used convolutional neural network (CNN) confronts similar challenges of view confusion and exhibits a limited capacity for global information aggregation due to its localized receptive fields [13].

Training strategies rooted in supervised learning (SL) demand a substantial volume of costly medical annotations to guide models in acquiring task-specific representation [14], which limits generality and flexibility in representation learning. Contrastive self-supervised learning (SSL) paradigms can acquire comprehensive representation based on the accumulated unlabeled data, which have gained substantial attention within physiological signal processing [15][16][17]. However, SSL assumes representation invariance to construct proxy tasks [18], a premise that often diverges from real-world scenarios. Besides, its sensitivity to data augmentation and batch size limits practical application [19]. Masked autoencoders (MAE) mitigate these challenges by masking the original signal and performing signal reconstruction [20]. Present SSL-based training strategies primarily target single-view signals, with a single pretraining task that may not perfectly align with diverse downstream tasks [21]. Consequently, current model architectures and training strategies seem inadequate in enabling effective MVF.

The complex missing-view scenarios in real-world applications pose significant challenges to the flexibility and robustness of MVF methods. An intuitive approach is to consider each missing-view scenario as a distinct task, requiring models capable of multitasking. However, managing numerous tasks demands frequent model training and increased storage capacity, which is impractical, particularly in today's increasingly complex model architectures. Resorting to conventional finetuning frameworks to reduce training costs may risk catastrophic forgetting resulting from iterative finetuning [22]. Inspired by the popular parameter-efficient finetune (PEF) technique applied in vision and language processing (VLP) models [23][24], we employ task-specific prompts to indicate different input distributions, which empower a well-trained MVF model with the flexible adaptation to complex missing-view scenarios.

Our detailed analysis revealed a negative response to two urgent questions mentioned earlier. This paper focuses on ECG and PPG signals. To mitigate the view confusion caused by channel concatenation operation, we advocate for the view-centric Transformer (VCT) as a backbone. VCT embeds sub-series from each view independently into a single-view token, concurrently equivalent to expanding the model's local receptive field [25]. The resultant tokens are inherently more view-centric and can be better leveraged by an MSA mechanism to aggregate cross-view and temporal information. To facilitate VCT in acquiring comprehensive representations, we propose the Multitask Masked Autoencoders (M2AE), encompassing contrastive ECG-PPG (CEP) learning and cross-view PPG and ECG reconstruction. Addressing intractable missing-view issues, we introduce two types of learnable missing-aware prompts to alleviate the performance degradation and the cumbersome finetuning procedures. Our contributions encompass four primary aspects:

1. **Straightforward yet effective MVF architecture:** the proposed VCT implements view-centric tokenization tailored for multi-view CVS data and leverages the classical MSA mechanism for capturing crucial cross-view and temporal interactions.

2. **Self-supervised pretraining strategy:** the proposed M2AE bolsters the joint understanding of ECG and PPG signals. Importantly, our findings indicate that supervision sources need not be confined solely to specific medical labels; different perspectives of CVS can effectively supervise each other.

3. **Detailed investigation about view missing:** to our knowledge, we first introduce the detailed missing-view scenario, where the view missing may vary across each data sample, either in the training or testing. Besides, we meticulously investigate the MVF model's robustness against view-incomplete.

4. **Flexible prompt technique:** we insightfully introduce the prompt technique, widely adopted in the VLP field, into the physiological signal field to address the general view-incomplete scenario. Additionally, we conduct a detailed investigation into the impact of prompt configuration variations, specifically examining the effects of the length and placement of two distinct prompt designs.

## 2. RELATED WORK

### 2.1 Network Architecture for Multi-View Cardiovascular Signals

Conventional practice treats multi-view cardiovascular signals as multivariate time series or multi-channel signals. CNN, which is adapted to process one-dimensional multi-channel signals, is commonly used to identify local pathological patterns in 12-lead ECGs for accurate cardiac arrhythmia classification [26][27][28]. However, CNNs struggle to model global information and inherent long-range dependencies in physiological signals [6]. Thus, studies have integrated Long Short-Term Memory (LSTM) layers after CNNs, achieving partial success in mitigating these limitations [29][30]. Recent studies have endeavored to replace the final layer in hybrid architectures with more robust multi-head attention layers, advancing global modeling in neural network structures [31][32]. This evolution extends to joint processing of ECG and PPG, where hybrid architectures—primarily using CNNs as the backbone supplemented with LSTM or MSA layers capturing temporal information—prove pivotal in tasks like blood pressure estimation and cardiac morbidity detection [3][33][34][35]. Nonetheless, the above methods, characterized by channel-wise concatenation, can lead to view-confusion issues. A small number of researchers have explored specific encoders for individual views, enabling effective MVF through carefully designed view fusion modules [5][36][37], thereby effectively alleviating view confusion issues. However, the increased parameter due to multiple encoders and the need for parallel processing of multiple views raise questions about the practical application of such methods. Hence, developing a straightforward yet efficient MVF model is critical

### 2.2 Self-Supervised Learning on Cardiovascular Signals

SSL has nearly become the standard practice in natural language processing (NLP) and computer vision, widely applied in various tasks [19] [38] [39]. In contrast, multitask SSL unravels intrinsic connections between different data modalities, such as images and texts [40][41]. Numerous efforts have focused on crafting specialized SSL methods for cardiovascular signals, in which approaches primarily revolve around two paradigms: instance discrimination and signal reconstruction. Well-established contrastive frameworks like SimCLR[18], BYOL[42], and SwAV[43], after adaptable modification for one-dimensional signals, have demonstrated success in handling cardiovascular signals like PPG, single-lead ECG, and 12-lead ECG [14][15][17]. To address individual and temporal variability inherent in cardiovascular signals, researchers have innovated specific data augmentation to meet the requirement of contrastive methods on subtle data augmentation and obtain representations with broader temporal and individual applicability [44][45]. Nevertheless, studies indicate that the optimal contrastive methods and augmentation strategies vary with different datasets [46]. The reconstruction task aims to reconstruct masked parts of input signals through the MAE employing Transformer architecture, thus minimizing sensitivity to data augmentation. A previous study applied sample points as the basic elements, resulting in elevated computational expenses [47]. While using sub-series as the basic element for synchronous temporal masking is prevalent in 12-lead ECG [10][48], it is not conducive to learning long-range dependencies across views in multi-view signals. Additionally, the above SSL techniques for cardiovascular signals often rely on a single proxy task, limiting their impact on enhancing downstream task performance.

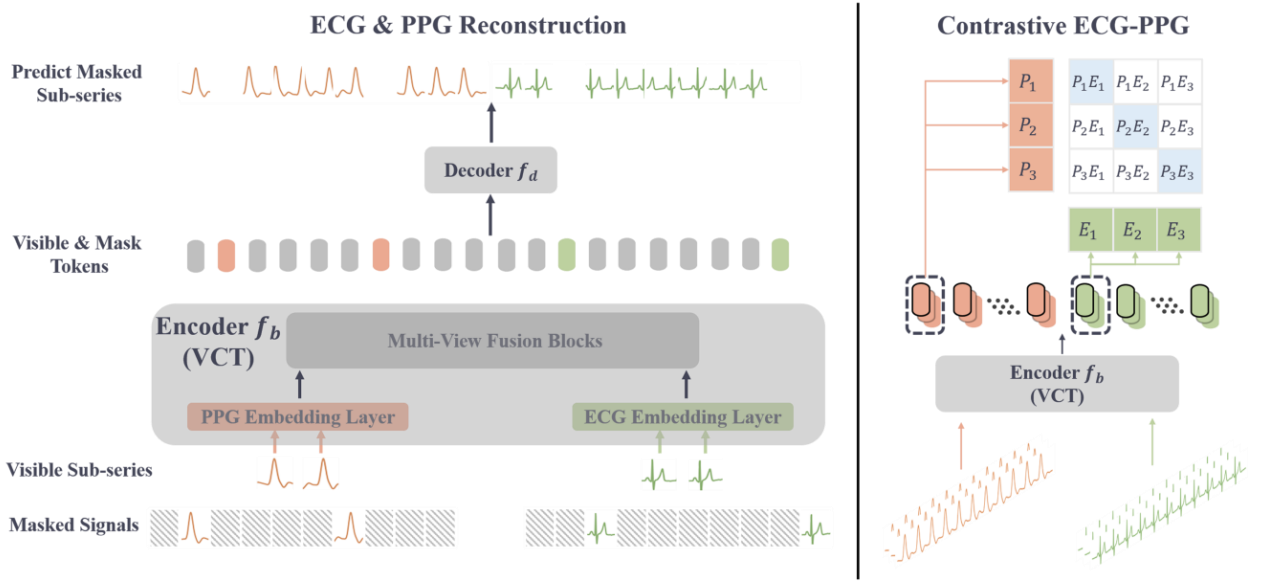

**Fig. 2.** Detailed illustration of the proposed VCT and M2AE. The masked ECG and PPG reconstruction is explained using a batch containing one sample, where the special learnable token for each view is omitted. The contrastive ECG-PPG learning is explained using a batch containing three samples.

### 2.3 Parameter-Efficient Finetune

Prompts have emerged as effective transfer learning and multitasking techniques in NLP and have found widespread use in VLP. Prompts involve modifying the input sequence to guide pretrained models in adapting to various downstream tasks [49]. Through manual prompts, pretrained models can demonstrate robust few-shot or zero-shot capabilities in these tasks [14]. Automatic prompt learning in continuous space, without human intervention, has recently garnered increased attention [23][24]. Adapters, another notable technique in PEF, modify or add parameters to specific layers within pretrained models [50][51]. These PEF technologies aim to employ minimal parameters, enabling heavy pretrained models to adapt to diverse downstream tasks flexibly.

## 3. METHODS

### 3.1 Problem Definition

**Multi-view Fusion:** A standardized nomenclature for signals simultaneously acquired from distinct body parts or sensors has not been established in cardiovascular signal analysis. While 'multimodal signals' are broadly acknowledged and employed [52][53], this paper opts for 'multi-view' to denote such signals due to the structural similarities across various physiological signals.

To be the simplest but without loss of generality, we only consider the two most common views M=2 in wearable devices, ECG and PPG ($m_1$ for ECG and $m_2$ for PPG). We adopt an SSL-based pretraining strategy instead of an SL approach for task-specific representation. We aim to fully utilize the physiological data easily collected by widely used sensors to craft comprehensive representation. Properly designing the proxy task proves crucial for an effective SSL method. The synchronized ECG-PPG pair, represented as $x^{m_1}$ and $x^{m_2}$, extracted from the multi-view CVS dataset $D^c$, is used to obtain multi-view representations via the following loss function:

$$\theta^*, \theta_1^*, ..., \theta_S^* = \arg\min_{\theta, \theta_1, ..., \theta_S} \sum_{s=1}^{S} L_s(Y_s, f_s^d(f^b(x^{m_1}, x^{m_2}; \theta); \theta_s)) \quad (1)$$

$S$ denotes the number of agent tasks; $Y_s$ is the fictitious supervision that depends on the design of the proxy task; Ls is the loss function associated with the proxy task; $f_s^d$ refers to the decoder along with its corresponding parameters $\theta_1, ..., \theta_s$, and $f^b$ stands for the backbone network and its corresponding parameter $\theta$. When the pretraining with multiple optimization objectives is completed, only $f^b$ is utilized for downstream tasks.

**View Missing:** regarding the common settings in missing-modality problem [23][54], we use the dataset $D$ to simulate complex missing-view scenarios in real applications, which can be decomposed into three subsets, i.e., $D = \{D^c, D^{m1}, D^{m2}\}$, where $D^c$ denotes the subset of datasets with complete views, and $D^{m1}$ and $D^{m2}$ denote the subset of datasets with incomplete views (i.e., ECG only and PPG only), respectively. To preserve the format of multi-view data, we assign pseudo-inputs $\tilde{x}^{m_1}, \tilde{x}^{m_2}$ (i.e., signal amplitude of 0) to the missing view data, resulting in the $\widetilde{D}^{m_1} = \{x_j^{m_1}, \tilde{x}_j^{m_2}, y_j\}, \widetilde{D}^{m_2} = \{\tilde{x}_k^{m_1}, x_k^{m_2}, y_k\}$. Consequently, the multi-view dataset incorporating the missing views can be reformulated as $\widetilde{D} = \{D_c, \widetilde{D}^{m_1}, \widetilde{D}^{m_2}\}$.

In employing the PEF technique, we plug a tiny amount of learnable parameters, denoted as $\theta_{pef}$, into the pretrained backbone network $f^b$. This approach necessitates updating solely $\theta_{pef}$ and task-specific head $f_t$. It circumvents the drawback inherent in the traditional finetuning methods, which typically require storing multiple sets of model parameters ($\theta$, $\theta_t$) when confronted with complex scenarios or downstream tasks. This strategy significantly enhances computational resource utilization, which is particularly crucial in the edge device. Furthermore, it contributes to adeptly handling general missing-view scenarios. The objective function is defined as follows:

$$\theta_l^*, \theta_t^* = \arg\min_{\theta_l, \theta_t} L_t(Y_t, f^t(f^b(x^{m_1}, x^{m_2}; \theta; \theta_{pef}); \theta_t)) \qquad (2)$$

$L_t$ and $Y_t$ denote the target of the downstream task and labels; $(x^{m_1}, x^{m_2})$ represents the samples from the missing-view dataset $D$.

### 3.2 View-Centric Transformer (VCT)

Our proposed VCT utilizes the simpler encoder-only architecture of Transformer [55], comprising two key components: the view-centric tokenizer for converting ECG and PPG signals into corresponding token sequences and the MVF module for extracting fusion information.

**View-centric tokenizer:** as shown in Fig. 2, since both tokenizers are structurally symmetric, and owing to the uniform format of input data from distinct views, the ECG tokenizer is illustrated as an example below. Initially, the ECG segment, with a length of $W$, is first segmented into $N$ sub-series $\{s_1, s_2, ..., s_N\}$, where $s_n \in \mathbb{R}^{P \times 1}$ and $P$ denotes the length of a sub-series, constituting a sequence characterizing the waveform amplitude across various time steps. Then, all sub-series undergo embedding into $d$-dimensional tokens through a linear layer $E^e \in \mathbb{R}^{P \times d}$, and a special learnable token $s_E$ is prepared to aggregate view-specific information. Finally, the positional encoding $E_{pos}^e \in \mathbb{R}^{(N+1) \times d}$ and the learnable view type tokens $E_{type}^e \in \mathbb{R}^{(N+1) \times d}$ are integrated into the ECG token sequences to inform the model of the temporal and view information, depicted as follows:

$$X^e = [s_E; s_1 E^e; s_2 E^e; ...; s_M E^e] + E_{pos}^e \qquad (3)$$
$$+ E_{type}^e$$

PPG characterization can be obtained by a similar process as described above:

$$X^p = [s_P; s_1 E^p; s_2 E^p; ...; s_M E^p] + E_{pos}^p \qquad (4)$$
$$+ E_{type}^p$$

Ultimately, a long sequence of multi-view CVS signals can be obtained by concatenation, where $h^0 \in \mathbb{R}^{(2N+2) \times d}$

$$h^0 = [X^e; X^p] \qquad (5)$$

**Multi-view fusion module:** $h^0$ Iterate through $L$ classical transformer layers to fuse information from distinct views across various time steps until the final token sequence $h^{L-l}$ is obtained. The pooling layer is a linear mapping layer that transforms the output sequence $h^{L-l}$ into the desired dimensions. The nonlinear activation function acts on the initial token of the sequence.

$$h^i = \text{MSA}\left(\text{LN}(h^{i-1})\right) + h^{i-1}, \qquad (6)$$
$$l = 0, ... L - 1$$

### 3.3 Multitask Masked Autoencoders (M2AE)

As illustrated in Fig. 2, our proposed M2AE encompasses three proxy tasks to facilitate the joint understanding of ECG and PPG for acquiring multi-view representation ($S = 3$ in Eq. (1)) for facilitating different downstream tasks. These tasks employ the mean squared error (MSE) loss function, which evaluates the reconstruction fidelity of ECG and PPG signals and the Info Noise Contrastive Estimation (InfoNCE) loss function designed to align ECG-PPG signal pairs.

**ECG/PPG Reconstruction:** Masked SSL has recently witnessed great success in NLP (e.g., BERT) and computer vision (e.g., MAE). However, the potential of MAE in multi-view signals, especially within physiological signals, remains largely unexplored. In the framework of a multimodal Transformer, various input modalities, despite potentially showing different data structures or physical meanings, are embedded into a formally consistent sequence of discrete tokens under the shared feature space, thereby enabling the fusing of heterogeneous data through the MSA mechanism. It holds promise for modeling multi-view signals using M2AE.

In Fig. 2, the input token sequences from different views undergo random masking with the masking rate $\alpha$. These sequences are then processed through the MVF module to capture the representations of observable sub-series. Higher $\alpha$ notably reduces the sequence length, lowering the computational overhead during pretraining. To represent the masked sub-series during the reconstruction phase, an additional learnable token is introduced, repeated following the number of masked sub-series, and combined with the visible tokens generated by the encoder to form the complete token sequence. Simultaneously, positional encoding is added to provide temporal information to the decoder, aiding in the precise reconstruction of corresponding sub-series across different time steps. This study aligns with He et al.'s approach, employing a uniform masking strategy to enhance overall performance. Moreover, the multi-view masking strategy obliges the model to investigate the temporal dependency within the timing signals and the interrelationships between distinct views when reconstructing the masked sub-series. Considering ECG and PPG signals' pseudo-periodic characteristics, a higher masking rate can help prevent the model from superficially memorizing signal patterns, emphasizing a deeper understanding of their intrinsic mechanisms.

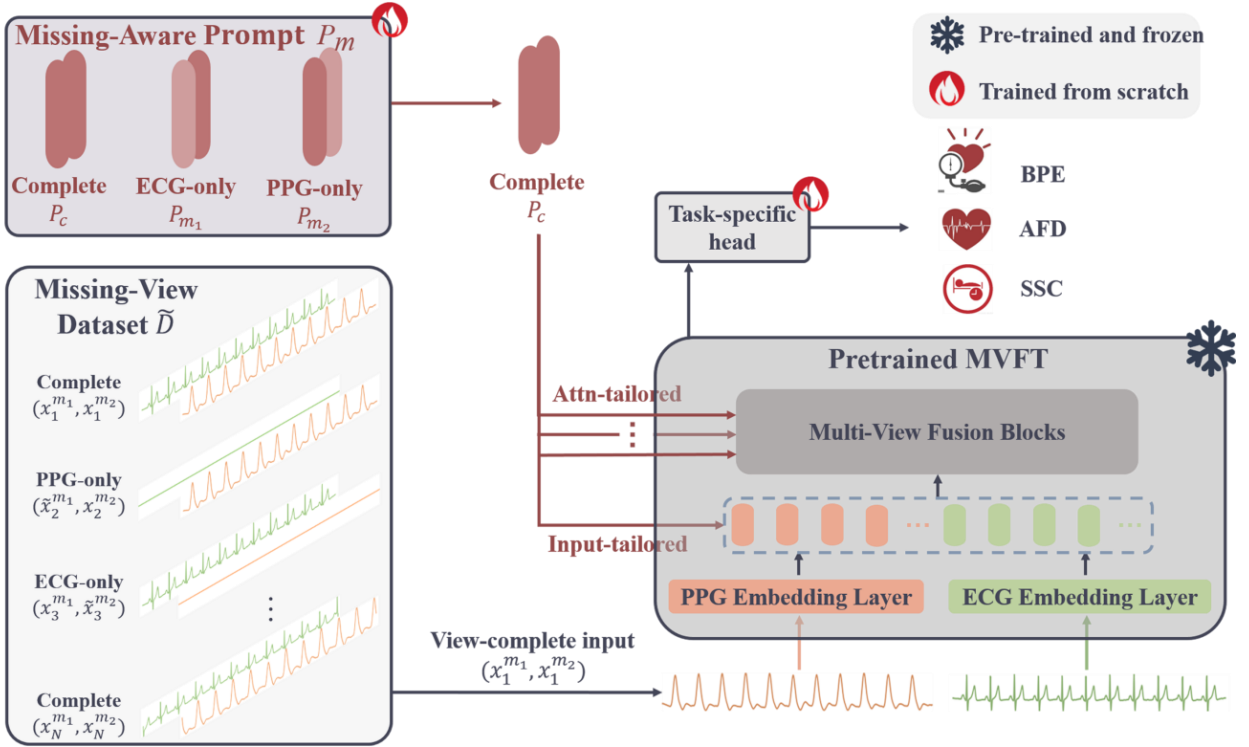

Fig. 3. An overview of our proposed missing-aware prompt. We first select the corresponding prompt $P_m$ based on the specific missing-view situation, where $\tilde{x}^{m_1}$ and $\tilde{x}^{m_2}$ is used for the corresponding view missing. Then, we design two methods to attach the prompt to multiple MSA layers. The features that integrate missing view information are provided to the task specific header for prediction. It should be noted that only task specific heads and missing-aware prompt need to be trained.

**Contrastive ECG-PPG (CEP):** ECG and PPG capture the cardiac electrical and cardiovascular mechanical activities, respectively, where the latter is influenced by the former, suggesting a consistent representation of cardiovascular information in both signals. Building on the insights from OpenAI's CLIP [14], we leverage the concept of mutual supervision between ECG and PPG to construct a proxy task. Specifically, synchronized ECG and PPG fragments from the identical person form positive sample pairs, while any dissatisfaction in synchronicity or identity categorizes instances as negative samples. Our methodology involves using VCT to concurrently encode both signals, followed by applying InfoNCE to evaluate their alignment within the feature spaces.

### 3.4 Prompt Learning for View Missing

Inspired by the prompt for missing modality in multimodal learning, we employ missing-aware prompts in guiding pretrained models conditioned on different input cases of missing view. Fig. 3 illustrates the operation procedure: initially, $M^2$ -1 prompts are prepared for the $M$-view task (three missing-aware prompts for the ECG-PPG data). These prompts are pretended to the input according to the type of missing view. Specifically, given a pretrained VCT $f^b$ comprising $L$ MSA layers, the missing-aware prompt $p_m{}^i \in \mathbb{R}^{N_p \times d}$ is consistently attached to the token sequence $h^i$ of respective $i$-th layer, where $N_p$ represents the prompt length, and $m \in \{c, m_1, m_2\}$ delineates different missing-view cases. The structure of the extended token sequence with each layer is as follows:

$$h_p^i = f_{prompt}(p_m^i, h^i) \qquad (7)$$

When implementing a downstream task with a view missing, optimization is exclusively directed towards the parameters $\theta_t$ of the task-specific head and the parameters $\theta_P$ associated with the missing-aware prompts $P_m$ spanning multiple layers. The overall objective function with trainable parameters is thus formulated.

$$\begin{aligned}\theta_p^*, \theta_t^* \\ = \arg\min_{\theta_p, \theta_t} L_t(Y_t, f^t(f^b(x^{m_1}, x^{m_2}; \theta_P); \theta_t))\end{aligned} \qquad (8)$$

which $(x^{m_1}, x^{m_2})$ comes from the dataset with an incomplete view $\tilde{D}$, $L_t$, $Y_t$ denote the downstream task's loss function and label, respectively.

Adding extra parameters to the pertrained model can improve performance, which is logically sound. Hence, the configuration and positioning of prompts are crucial for fully leveraging the potential of prompt learning. The following paragraph will thoroughly introduce the design of function $f_{prompt}$ that attaches prompts to each chosen MSA layer. While common prompt learning methods generally add prompts to the input sequence to guide downstream task predictions, this paper uniquely focuses on the challenge of missing view. We have adopted two prompt conFigurations further to augment the performance and adaptability of prompt learning.

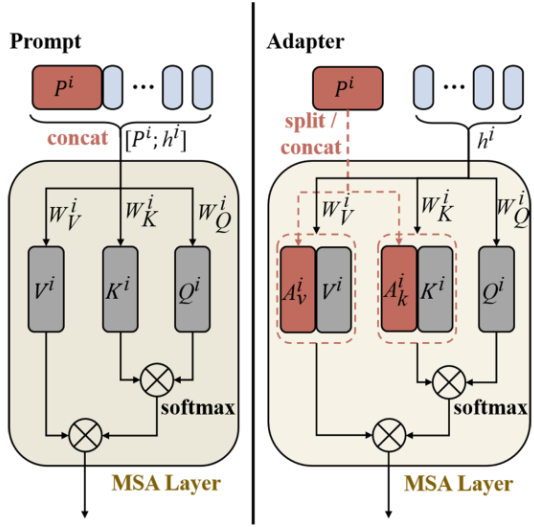

Fig. 4. An overview of our proposed missing-aware prompt. We first select the corresponding prompt $P_m$ based on the specific missing-view situation, where $\tilde{x}^{m_1}$ and $\tilde{x}^{m_2}$ is used for the corresponding view missing. Then, we design two methods to attach the prompt to multiple MSA layers. The features that integrate missing-view information are provided to the task specific header for prediction. It should be noted that only task specific heads and missing-aware prompt need to be trained.

**Input-tailored approach:** We adopt the widely used prompt design, directly attaching prompts to each layer's input sequence for concatenation operations. Thus, the prompt function can be expressed as:

$$f_{prompt}^{input}(p_m^i, h^i) = [p_m^i; h^i] \qquad (9)$$

Assuming a series of consecutive $L_p$ MSA layers, where the input sequences are concatenated with prompts of length $N_p$, the length of the resulting sequence from the whole model becomes $(N_p * L_p + 2 * N + 2)$. This setup ensures the interaction of prompts in the current layer with tokens from previous layers, preventing the loss of missing-view information at greater depth. However, this also leads to a gradual increase in the proportion of prompts within the overall token sequence, which may be less favorable for certain downstream datasets.

**Attention-tailored approach:** This approach implements prompt by modifying the input of the MSA layer. As shown in the right panel of Fig. 4, the missing-aware prompt is decomposed into two sub-prompts, $p_k^i$ and $p_v^i$ of length $L_p$ /2. These sub-prompts are concatenated with the keys and values in the MSA. Then, the classical attention operation can be reformulated as follows:

$$f_{prompt}^{attn}(p_m^i, h^i) = Attn^i(p_m^i; h^i) \qquad (10)$$

$$Attn^i = \text{softmax}\left(\frac{Q^i[p_k^i, K^i]^T}{\sqrt{d}}\right)[p_v^i; V^i] \qquad (11)$$

The Attention-tailored approach offers an alternative strategy for modifying the MSA layer's internal mechanism to guide the pretrained model in alleviating the missing-view problem. Importantly, it maintains the length of the final output token sequence.

## 4. Experimental Setup

### 4.1 Datasets and Metrics

#### 4.1.1 Datasets

We have chosen three key health monitoring tasks, each paired with the ECG and PPG data dataset, to evaluate the effectiveness of our proposed method.

**Blood Pressure Estimation (BPE):** Wang et al. took a rigorous signal quality assessment to clean two publically-available databases (e.g., Multi-parameter Intelligent Monitoring for Intensive Care (MIMIC) and VitalDB [56]. This effort culminated in creating PulseDB, the most extensive curated dataset in this field, specifically tailored for benchmarking BPE models. Comprising ten meticulously defined subsets, PulseDB features 10-second non-overlapping segments of synchronized ECG, PPG, ABP, diastolic blood pressure (DBP) and systolic blood pressure (SBP), collected from 5361 subjects from MIMIC and VitalDB.

**Atrial Fibrillation Detection (AFD):** Some existing datasets utilized for AF (Atrial Fibrillation) detection, primarily based on single-view physiological signals (PPG or ECG), fail to fulfill the requirements for multi-view physiological signals in **this** study. Additionally, discrepancies in sampling rates between PPG and ECG signals within the AF detection dataset, alongside unaligned acquisition times, hamper the feasibility and analysis of multi-view experiments. The PERformAF dataset, curated by Charlton et al., comprises 20-minute data segments from 19 patients diagnosed with atrial fibrillation and 16 patients without atrial fibrillation [57], extracted from the MIMIC dataset.

**Sleep Stage Classification (SSC):** We randomly extracted data from 800 subjects from the representative sleep dataset, Multi-ethnic study of atherosclerosis (MESA) [58][59]. The dataset adheres to the guidelines outlined by the American Academy of Sleep Medicine (AASM), delineating sleep stages into wakefulness, REM, and three non-REM stages: N1, N2, and N3. For this study, our focus centered on the four common sleep staging categories: wakefulness, mild (N1/N2), depth, and REM sleep.

#### 4.1.2 Preprocessing

The largest subset, 'Train_Subset' within the PulseDB, served as the basis for pre-training our proposed VCT, which underwent evaluation for two downstream tasks: BPE and AFD. For the BPE task, we employed the additional 'VitalDB_Train_Subset' and 'VitalDB_CalBased_Test_Subset' subsets from the PulseDB for finetuning and testing. As our study did not consider models tailored to different input lengths, we processed the PERformAF dataset for the AFD task by segmenting each patient's data into 10-second windows (matching the PulseDB dataset) with an overlap rate set at 0.8 for data expansion. Subsequently, data allocation among training, validation, and test sets follows a 6:2:2 ratio, ensuring a balanced distribution.

We adhere to common AASM criteria for the MESA dataset by segmenting each patient's data record into non-overlapping 30-second segments [60]. To maintain consistency

| Task | AFD (Classification) | BPE (Regression) | SSC (Classification) |
|---|---|---|---|
| Dataset | PERFormAF | PulseDB | MESA |
| Sample Size | 0.03M | 5.17M | 6.31M |
| Label Type | AF / non-AF | DBP & SBP | Wake/Light/Deep/REM |
| Label Distribution | 0.012M / 0.014M | - | 2.7M / 2.6M / 0.6M / 0.4M |
| Input Length (Secs) | 10 | 10 | 30 |
| Sampling Rate (Hz) | 125 | 125 | 100 |

in the length of the two inputs, we resample both ECG and PPG signals to 100Hz. The resultant dataset is divided into training, validation, and test sets using the 6:2:2 ratios. We utilized the entire training set for pretraining and finetuning purposes. A detailed discussion about the effect of dataset proportion on the model performance can be found in subsection 5.2.

### 4.1.3 Evaluation Metrics

In BPE tasks, the conventional evaluation metrics often involve using the mean and standard deviation of estimation errors for DBP and SBP. However, we employed root mean square error (RMSE), a widely recognized overall indicator, to assess BPE estimation performance for clarity and ease of analysis. The accuracy serves as the evaluation metric for the AFD task. Notably, Table I reveals a significant category imbalance in the MESA dataset. Therefore, for the four-class classification problems, we utilized the F1-Macro score as the evaluation metric.

### 4.2 Model Settings

We standardize the VCT parameters across all datasets and training phases. Input ECG and PPG segments are uniformly divided into sub-series of length 50. A consistent embedding size of 512 is employed across the entire encoder. To ensure that VCT captures temporal information, we utilize cosine-based positional encoding. The encoder depth is set to 8, incorporating the classical MSA module with eight heads.

During the pretraining phase, we incorporate a lightweight decoder alongside the encoder to extract contextual information from obscured locations efficiently and reconstruct the invisible sub-series while minimizing computational overhead. Specifically, the depth and embedding dimension of the decoder are set to 3 and 256, respectively, with six heads in the MSA module. The decoder originally utilized for proxy tasks is replaced with a task-specific head, enabling the implementation of various downstream tasks.

### 4.3 Training Details

**Pretraining:** The VCT is pretrained on 10-second and 30-second inputs extracted from the PulseDB and MESA datasets, respectively, leveraging the proposed M2AE architecture for 100 epochs. The training utilizes the AdamW optimizer with a base learning rate of 1e-3 and a weight decay 0.05. We initiated the training with a warm-up phase comprising 10% of the total training epochs, followed by a cosine decay approach to diminish the learning rate to 0. Employing a batch size 2048, we trained the models across 2 A100 GPUs utilizing hybrid precision techniques. In the case of the model pretrained on

PulseDB, a mask rate ($\alpha$) of 0.8 was applied to each view, aiming to expedite pretraining and facilitate optimal representation learning. Contrastingly, considering the substantial motion artifacts within the MESA dataset, we set the masking rate $\alpha$ to 0.5. This choice aimed to mitigate the adverse impact of noise on the reconstruction task.

**Finetuning:** we utilize two common methods to evaluate the effectiveness of the MVF representation: linear probe and finetuning. The former involves freezing the weights of the pre-trained encoder and optimizing the added linear heads for downstream tasks. Conversely, the latter approach involves optimizing the entire model without weight freezing. Comparatively, the linear probe indicates the pretrained model's generality, equivalent to selecting pretrained representations using solely the linear layer. In this context, 'FineLast' and 'FineAll' refer to these two evaluation methodologies. The optimizer settings during finetuning are the same as the pretraining, except that the weight decay is 2e-2. Given the substantial dataset sizes in the BPE and SSC downstream tasks, we set the batch size to 1024 to expedite the finetuning process. In contrast, considering the smaller size of the AFD dataset, a batch size of 128 was utilized. Employing the early stopping method in our experiments is to prevent potential overfitting of the model

### 4.4 Settings of View Missing

It is crucial to highlight that our adoption of varying types and proportions of view missing is solely for quantitative analysis rather than underestimating the complexity of missing-view scenarios prevalent in real-world applications. We deliberately adopt a more aggressive setting where view missing is assumed to potentially affect every sample, both during the training and testing phases. For the ECG-PPG pairing downstream task, view missing manifests in three cases: missing ECG, missing PPG, or missing both. The view missing rate, $\beta\%$, denotes the proportion of samples with missing views in the entire dataset. In the case of missing ECG with $\beta\%$, it implies $\beta\%$ PPG-only data and $(1-\beta)\%$ complete data, conversely for the missing-PPG scenario. For the missing-both case, there are $\beta/2\%$ ECG-only data, $\beta/2\%$ PPG-only data, and $(1-\beta)\%$ complete data. The length of missing-aware is set to 20, selectively applied to layers 1 to 6 of the encoder. Its optimization strategy mirrors the finetuning approach established for downstream tasks in Section 4.3. By default, the missing rate $\beta\%$ is set to 70% in our experiments. The bigger the $\beta$, the severity of the view missing.

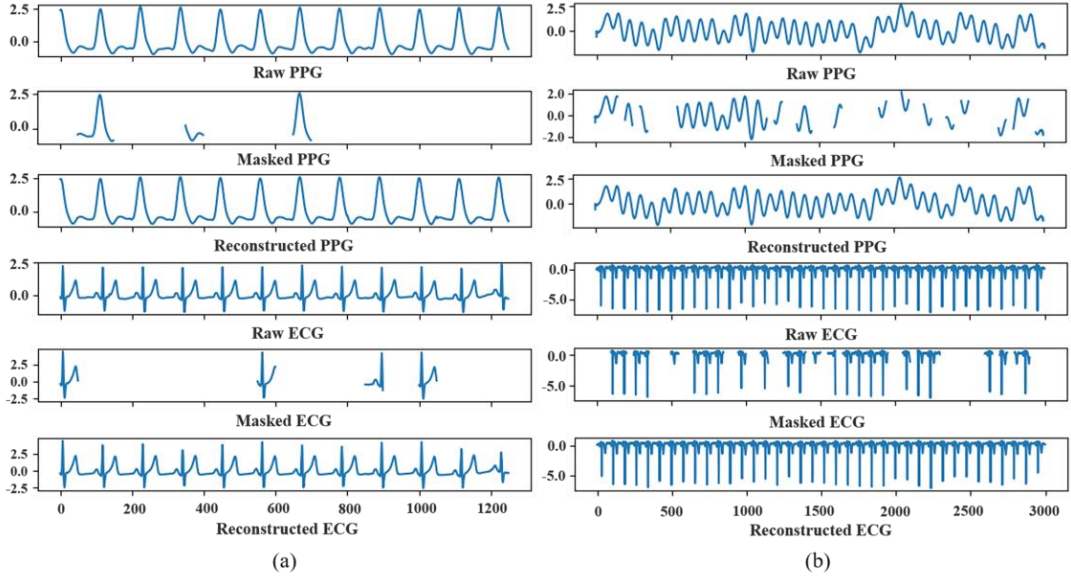

**Fig. 5.** The cross-view signal reconstruction result. (a) PulseDB dataset; (b) MESA dataset

Table II.
Comparison between different fusion strategies

| Fusion Strategy | Model | Train | Epoch | Param (M) | BPE (RMSE ↓) | AFD (ACC ↑) | SSC (F1-Marco ↑) |
|---|---|---|---|---|---|---|---|
| Single-View | Trans(PGG) | RandInit | 100 | 25.6 | 11.59 | 66.21 | 34.94 |
| | Trans(ECG) | RandInit | 100 | 25.6 | 12.32 | 96.77 | 30.50 |
| Multi-View | ResNetAttn | RandInit | 100 | 30.1 | 7.86 | 62.79 | 59.30 |
| | Trans | RandInit | 100 | 25.6 | 6.36 | 48.88 | 60.90 |
| Multi-View | VCT | RandInit | 100 | 25.9 | 4.09 | 58.16 | 73.73 |
| | VCT | FineLast | 20 | 25.9 | 10.36 | 88.13 | 45.00 |
| | VCT | FineAll | 20 | 25.9 | 3.47 | 97.99 | 80.85 |

## 5. Results and Discussion

### 5.1 Analysis of Optimal Fusion Strategy

We compared mixed-view and our proposed view-centric strategies to ascertain the optimal fusion approach. Mixed-view methodologies typically treat multi-view signals as multi-channel signals, amalgamating multi-channel sub-series from the same time step into a unified representation, consequently fostering view-confusion issues. Following the prevalent preference in physiological signal processing for hybrid architectures such as CNN-Transformer, we replaced the final pooling layer of the classic ResNet with an MSA layer [14]. In parallel, we replicated a Transformer model utilizing the mixed-view tokenization, as depicted in Fig. 1, for comparative experiments to explore the efficacy of view-centric tokenization. Furthermore, we tailored a Transformer to handle single-view signals, confirming the indispensability and benefits of MVF.

The results presented in Table II highlight the limitations of single-view signals for complex tasks like BPE and SSC. Notably, integrating an additional view's signal in the channel dimension significantly enhances the performance of the hybrid ResNetAttn and Transformer in MVF. Furthermore, comparing the performance of ResNetAttn and Transformer on the PulseDB and MESA datasets with rich samples indicates that the MSA mechanism, the cornerstone of Transformer, is more effective in global temporal modeling. However, due to its lack of inductive bias, a larger sample size is necessary compared to CNN to counteract this limitation. Our proposed VCT fusion methodology, rooted in view-centric tokenization for deeper exploration of effective temporal modeling and rational multi-view correlating, proves stronger than mixed-view methods.

Notably, single views exhibit strong performance in AFD tasks, primarily due to the detailed information from ECG signals concerning AF. Consequently, the primary view can adeptly handle detection tasks, with marginal gains from modeling cross-view interactions. Nevertheless, for handling intricate or noisy datasets, models necessitate the capability to encode cross-view relationships. Particularly in datasets reliant on two views for prediction, our approach presents a substantial advantage for temporal and cross-view interactions.

### 5.2 Analysis of Pretraining Paradigm

**Reconstruction visualization:** Fig. 5 shows the proxy task implemented by the VCT on the cleaned PulseDB and the noisy MESA dataset. Despite the lack of temporal synchronicity among visible sub-series of multi-view signals, our method adeptly reconstructs the shape and positioning of QRS waves (ECG signals) and peaks (PPG signals). This observation substantiates the method's ability to understand many potential waveform patterns proficiently. Furthermore, when applied to the more corrupted data within the MESA dataset, the method

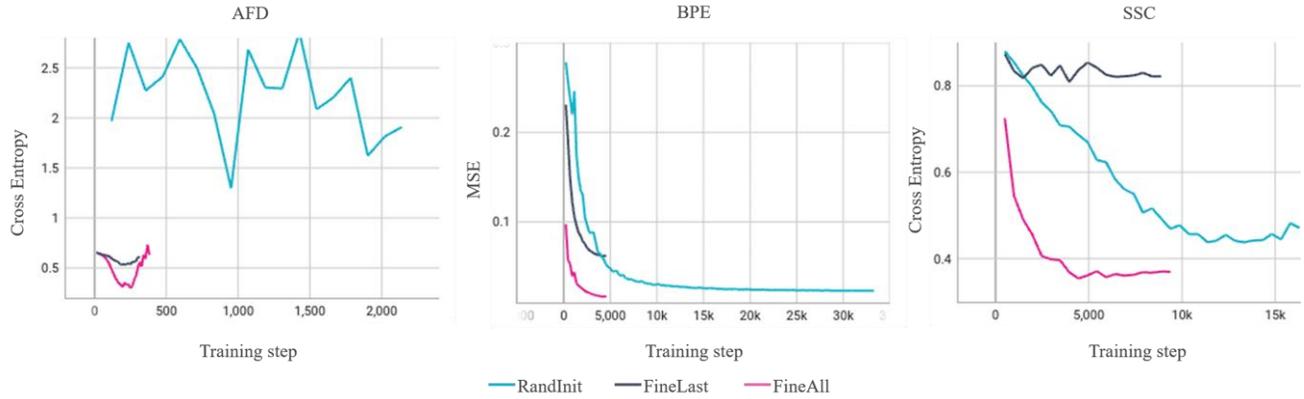

Fig. 6. Loss curve of VCT using different training strategies on validation set of three downstream tasks.

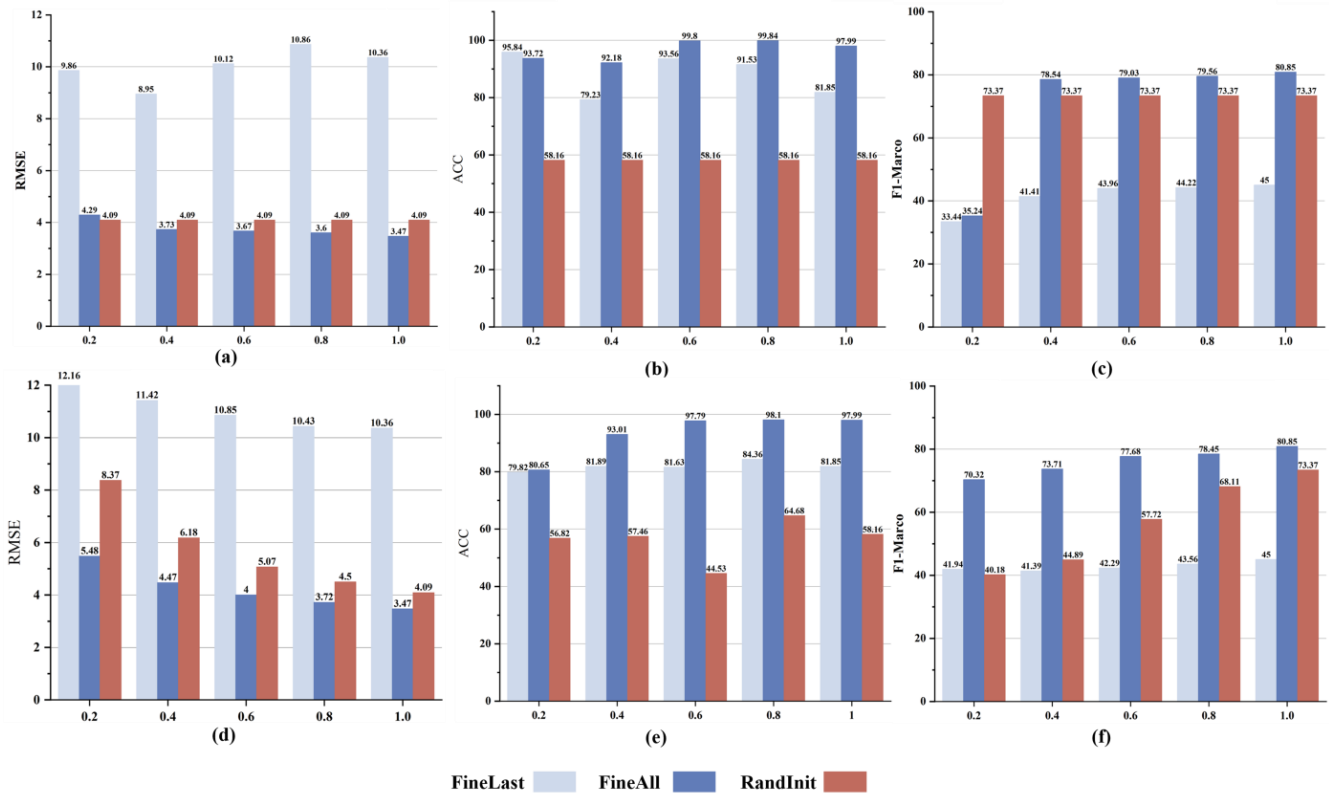

Fig. 7. Performance of VCT using different dataset sizes and training methods on three downstream datasets. (a), (b), and (c) represent different proportions of pretraining set over PulseDB, PERFormAF, and MESA, respectively; (d), (e), and (f) represent different proportions of training set over PulseDB, PERFormAF, and MESA, respectively

displays an impressive capacity to recover the underlying baseline drift inherent in the signal. This remarkable achievement indicates that the VCT via pretraining strategy based on M2AE has successfully encoded both global and local information within the multi-view CVS signals and holds promising potential for enhancing performance in subsequent downstream tasks.

**Downstream task:** Table II presents detailed experimental results from different training methods. Notably, the performance evaluation based on the FineAll method showcases more promising outcomes than the FineLast and RandInit methods. This difference is rational, given that the FineAll method forces the entire model's weights to suit the target task. Conversely, the FineLast approach freezes the entire model's encoder except for the task-specific head, precluding the injection of new knowledge. In the case of random initialization, the model has no prior knowledge to assist the learning process.

Fig. 6 illustrates the loss curves across three distinct training methods on three downstream validation sets. Comparative analysis reveals that the FineAll and FineLast methods exhibit lower initial training losses and more rapid loss reduction during the early stages of training compared to the random initialization method. Employing the FineAll method for 20 training epochs significantly yields superior results compared to 100 epochs utilizing random initialization. These outcomes fully demonstrate the powerful representation learning ability of our proposed M2AE. We attribute the above achievement to the simple philosophy of recovering masked sub-series based on visible sub-series, which does not rely on carefully constructed positive and negative samples.

Our proposed M2AE showcases significant potential in acquiring comprehensive representations while upholding signal length and sampling rate consistency. For both BPE and AFD tasks, we adopt the same subset of PulseDB for pretraining and different downstream datasets for evaluating, inherently adopting a one-to-many evaluation method. Table II shows that the findings demonstrate the efficacy of applying solely the pretrained features without additional knowledge injection (FineLast), achieving an AFD effect of 81.85 on the external PEFormAF dataset. Additionally, we conducted SL-based pretraining by utilizing BP label information from PulseDB and subsequently transferred this knowledge to the AFD task. The resultant accuracy using the FineLast and FineAll methods were 52.61 and 56.82, respectively. The significant discrepancy between SL-based and our proposed SSL-based approach emphasizes the latter's adeptness in circumventing task bias, facilitating the acquisition of generalized representations. Additionally, the low initial cross-entropy loss observed during finetuning on the validation set proved the possibility of the emergence of generalized models within physiological signal processing.

Significantly, we find that the scale of the downstream dataset impacts the effectiveness of finetuning. The PERFormAF dataset, distinguished by its notably constrained data volume, proves insufficient to support training the VCT from scratch. This underscores the significance of utilizing a straightforward and effective SSL-based pretraining method. Moreover, the irregular fluctuations witnessed in the validation set curve of the AFD task suggest the potential for overfitting on a small dataset containing merely 0.03M samples. This alignment with reference [61] provides a theoretical basis for applying early stopping techniques.

### 5.3 Performance on the Missing-View Data

This section focuses on analyzing the efficacy of the missing-aware prompt in handling intricate missing-view scenarios, alongside evaluating the efficiency of integrated learnable prompts. We followed the baseline setting for comparison commonly used in prompt learning and incomplete data [23]. The main baseline for comparison is the model based on a linear probe, where the performance gained by the proposed approach concerning such baseline directly reflects the benefits of the missing-aware prompt.

**View Dominance:** Table III presents a comprehensive overview of results about various missing-view scenarios, all with a 70% missing rate, across the three downstream tasks. Severe view missing notably degrade baseline performance, yet the inconsistent negative effects across diverse scenarios at the same rate arise due to distinct view dominances across different tasks. In the cardiovascular context, BP delineates the pressure exerted by blood on vessel walls during flow, aligning closely with the PPG signal that measures changes in blood volume within vessels. Conversely, AF signifies the disruption in regular atrial electrical activity, replaced by rapid and disorderly fibrillation waves, a characteristic efficiently captured by ECG. Therefore, among different combinations of specific view missing, the absence of a corresponding dominant signal profoundly impacts both downstream tasks. Regarding the Sleep Staging Classification (SSC) task, prior studies emphasize the efficacy of leveraging either ECG or PPG solely for achieving comparable performance [62], signifying an equivalent importance of ECG and PPG dominance.

**Comparison between different prompts:** Table III also presents quantitative results for employing two distinct missing-aware prompts all with a 70% missing rate. By examining Table III alongside Fig. 8-9, several observations become apparent. Primarily, across the three downstream datasets, a consistent trend emerges irrespective of the severity of view missing and the combinations of missing views: an incremental decline in overall performance as the severity of view missing increases. However, both prompting methods significantly enhance the model's ability to handle view missing. This underscores the effectiveness of our method in mitigating missing-view problems, notably reflected in the reduced RMSE for the BPE task and increased ACC and F1 for the AFD and SSC tasks. Secondly, while the input-tailored prompt generally exhibits greater performance improvement compared to the

Table III.

Quantitative results on the BPE, AFD, and SSC tasks with missing rate $\beta\% = 70\%$ under various missing-view scenarios. Bold number indicates the best performance. Attn-tailored indicate attention-tailored missing-aware prompt.

| Datasets | Missing rate | Train & Test | | Baseline | Attn-tailored ($\triangle\uparrow$) | Input-tailored ($\triangle\uparrow$) |
|---|---|---|---|---|---|---|
| | | PPG | ECG | | | |
| BPE (RMSE) | 70% | 100% | 30% | 12.62 | 11.68 (7.5%) | **10.71 (15.1%)** |
| | | 30% | 100% | 13.43 | 11.55 (14.0%) | **9.69 (27.9%)** |
| | | 65% | 65% | 13.18 | 11.89 (9.8%) | **10.36 (21.4%)** |
| AFD (ACC) | 70% | 100% | 30% | 69.29 | 71.66 (3.4%) | **77.07 (11.2%)** |
| | | 30% | 100% | 75.30 | 79.75 (5.9%) | **81.63 (8.4%)** |
| | | 65% | 65% | 62.72 | **74.61 (19.0%)** | 67.77 (8.1%) |
| SSC (F1-Marco) | 70% | 100% | 30% | 39.34 | 42.22 (7.3%) | **48.35 (22.9%)** |
| | | 30% | 100% | 38.16 | 41.61 (9.0%) | **46.75 (22.5%)** |
| | | 65% | 65% | 37.23 | 40.72 (9.4%) | **46.53 (25.0%)** |

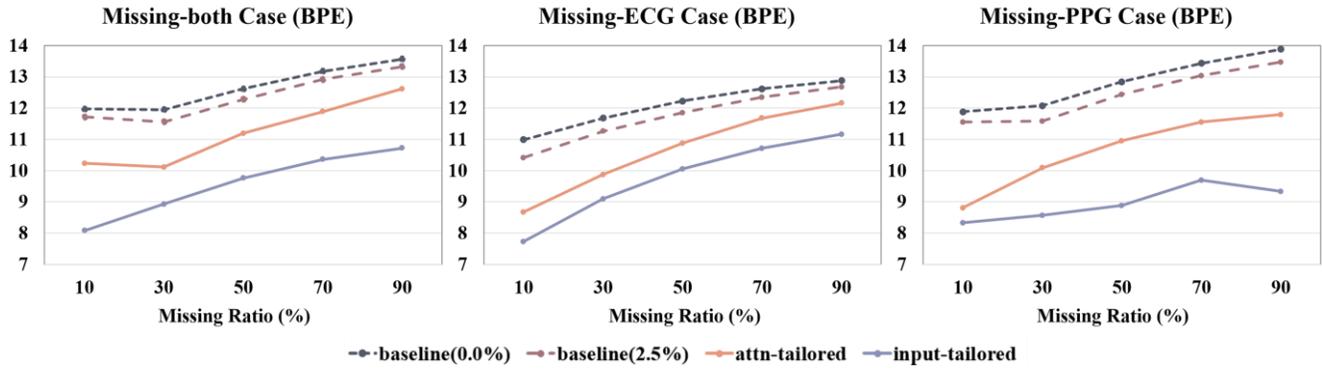

Fig. 8. Quantitative results on the PulseDB with different missing rates under different missing-modality scenarios. Each data point represents that training and testing are with the same $\beta$% missing rate.

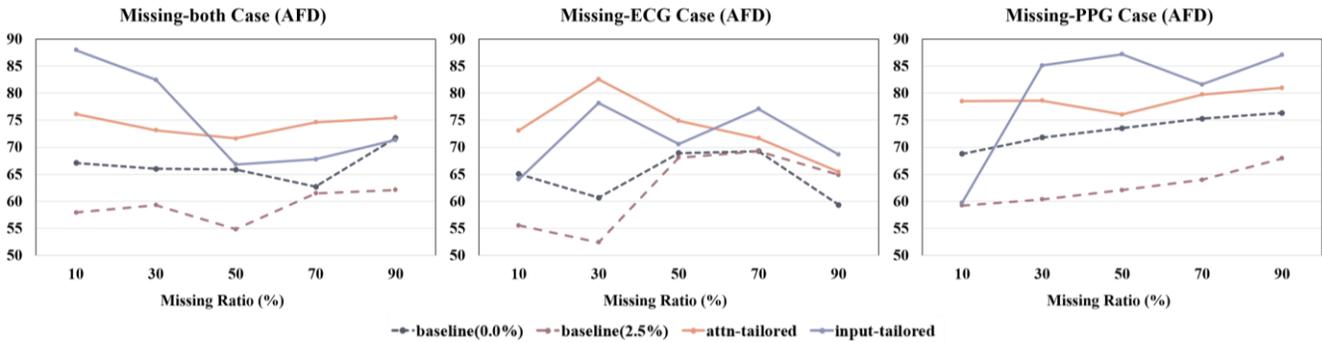

Fig. 9. Quantitative results on the PERFormAF with different missing rates under different missing-modality scenarios. Each data point represents that training and testing are with the same $\beta$% missing rate.

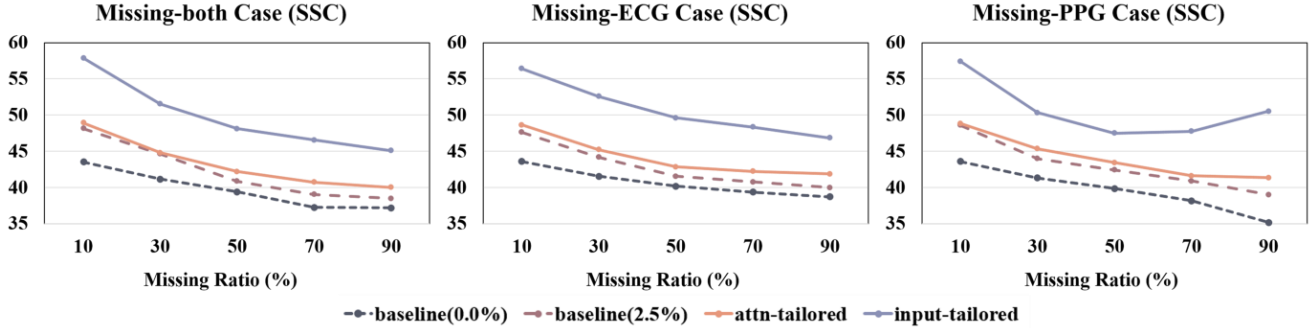

Fig. 10. Quantitative results on the MESA with different missing rates under different missing-modality scenarios. Each data point represents that training and testing are with the same $\beta$% missing rate.

attention-tailored prompt, its stability, particularly on the PERFormAF dataset, is comparatively lower than that of the attention method. Both methods provide different conFigurations of prompts to guide the model prediction depending on the view missing.

As detailed in Section 3.4, input-tailored methods inherit the prompt information from proceeding layers, suggesting a progressive accumulation of prompts during the model's forward propagation. When datasets are limited and prompt sequences are extensive, this behavior might introduce ambiguity, impeding the model's ability to learn task-specific features. This becomes particularly evident in scenarios where the input multi-view token sequence length stands at 50, while the prompt sequence length is set to 20, resulting in inferior performance of the input-tailored approach on the smaller

PERFormAF dataset. Conversely, the attention-tailored prompts are attached to the keys and values within selected MSA layers, only guiding the token sequence's adaptation within the current layer. While the input-tailored approach excelled on the other two datasets, the attention-tailored approach exhibits greater stability to dataset variations.

**Efficiency on Parameters:** Finetuning the entire pretrained VCT, consisting of 25.9M parameters, yields an RMSE metric 8.86 on missing-both data with a 70% missing rate from PulseDB. Conversely, our approach freezes the pretrained VCT and solely trains the missing-aware prompt containing 676K parameters (2.5% of the entire model), achieving a competitive RMSE metric of 10.47. The intuitive assumption that introducing additional parameters invariably enhances performance led us to conduct a comparative experiment. Adding a nonlinear layer of approximately 676K parameters to the task-specific header (baseline (2.5%)) aimed to rule out the above idea as a misconception of our approach. As shown in Fig. 8 and Fig. 10, the additional parameter (baseline(2.5%)) does not consistently enhance performance across large PulseDB and MESA datasets; instead, it leads to a significant decrease in performance on the smaller PERFormAF dataset. This highlights the crucial significance of configuring additional parameters effectively. It is noteworthy that we include the task-specific head in the trainable parameters. Thus, strictly speaking, parameters arising from the missing-aware prompt would be significantly less than the reported 2.5% in this study.

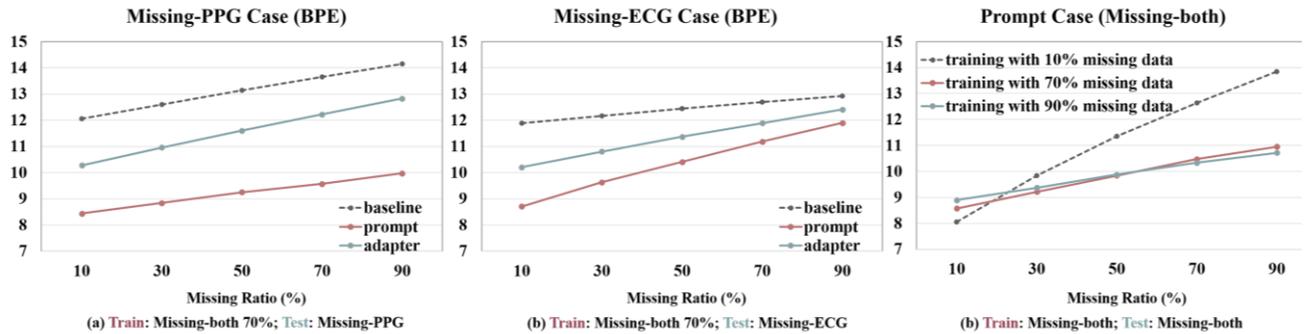

Fig. 11. Ablation study on robustness to the testing missing rate in different scenarios on PulseDB. (a) All models are trained on missing-both case with 70% missing rate, and evaluated on missing-PPG case with different missing rates. (b) All models are trained on missing-both case with 70% missing rate, and evaluated on missing-ECG case with different missing rates. (c) Input-tailored prompts are trained on missing-both cases with 10%, 70%, and 90% missing rate, which represents more view-complete data, balanced data, and less view-complete data, respectively. Evaluation is on missing-both case with different missing rates.

### 5.4 Ablation Study

**Effects of Different Proxy Tasks:** Our proposed M2AE incorporates ECG and PPG signal reconstructions and CEP, pretraining the VCT on unlabeled ECG-PPG pairs. The shared token masking rate $\alpha$ controls the impact of both reconstructions. We conduct detailed experiments within this subsection to ensure analytical simplicity and clarity using the clean PulseDB dataset. We rigorously adopt a FineLast approach to evaluate representation quality learned from varying combinations of mask rates and CEP tasks. Table IV demonstrates that higher $\alpha$ correlates with a gradual reduction in RMSE, implying that employing a higher mask rate aids the model in obtaining comprehensive multi-view representation. Furthermore, even the reconstruction task with the least effectiveness ($\alpha = 0.2$) outperforms using CEP alone as a proxy task. Nonetheless, CEP consistently reduces RMSE on the top of reconstruction tasks with different masking rates, indicating its role as a contrastive learning mechanism that captures high-level concepts shared among different views and facilitates MVF. It is a valuable auxiliary task that significantly improves representation learning in conjunction with the primary task.

**Robustness to missing rate:** we conduct further experiments to assess the robustness of the proposed missing-aware prompt across varying missing-view rates in the training and testing phases. Initially, models trained on the 'missing-both' case with a 70% missing rate are evaluated on the testing set encompassing different missing-view scenarios. These scenarios are represented in the left and middle panels of Fig. 11, showcasing missing-PPG and missing-ECG data. The prompt methods consistently outperformed the baseline model across diverse missing rates during the testing phase, regardless of the specific missing-view scenario. Furthermore, the input-tailored approach consistently exhibits a clear advantage over the attention-tailored approach. This trend suggests that the input-tailored method learns better on view-complete data and exhibits greater robustness against view-incomplete data.

Table IV.
Ablation study of M2AE based on FineLast method and PulseDB. Bold number indicates the best performance.

| CEP | $\alpha=0.2$ | $\alpha=0.5$ | $\alpha=0.8$ | RMSE |
|---|---|---|---|---|
| √ | | | | 15.44 |
| | √ | | | 14.49 |
| | | √ | | 13.04 |
| | | | √ | 11.67 |
| √ | √ | | | 13.17 |
| √ | | √ | | 12.24 |
| √ | | | √ | **10.36** |

The right panel of Fig. 11 shows the results of the input-tailored method trained with missing-both rates: 10%, 70%, and 90%. These rates represent different degrees of missing-view scenarios during training from more view-complete data (10% missing-both), balanced data (30% complete, 35% missing-PPG, and 35% missing-ECG), to severely missing-view data (90% missing-both). Notably, the performance curves corresponding to the 10% missing-both scenarios display significant fluctuations when tested across varying missing rates. It may stem from the under-optimization of the learnable missing-aware prompt, specifically targeting missing-view data, owing to the less missing-view data available for training. Conversely, models trained with balanced and severely missing-view data showcase almost indistinguishable performances, demonstrating stable and consistent trends across test situations. Interestingly, even the model trained with 90% incomplete data exhibits substantial competitiveness, particularly in testing on severely missing-view data. These above observations demonstrate the robustness of our missing-aware module in handling complex missing-view scenarios.

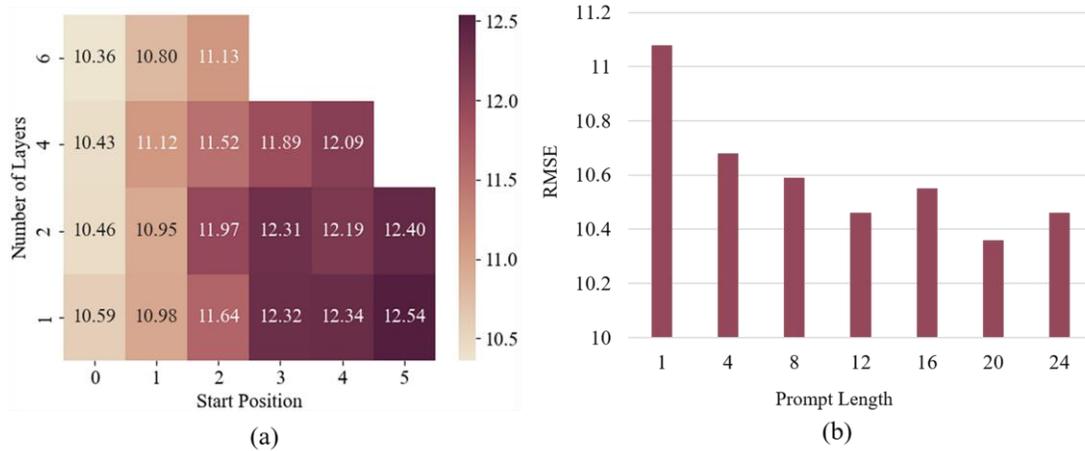

**Fig. 12.** Ablation study on the conFiguration of input-tailored prompts. (a) The effect of insertion location; (b) The effect of prompt length

**The effect of selected layers:** Fig. 12(a) illustrates the impact of the placement of input-tailored missing-aware prompts. The visualization portrays a discernible trend where performance consistently improves with the progressive attachment of prompts across multiple MSA layers. While the more critical factor is determining where to start attaching the prompts layer by layer, the results answer that prompting the model at an early stage obtains better results. One plausible explanation lies in the model's initial fusion of multi-view inputs, where deeper layers imply increased fusion levels. Therefore, guiding early layers by the missing-aware prompts is more effective, preserving the distinct characteristics of each modality before potential dilution during the fusion process.

**The effect of prompt length:** Fig. 12(b) shows the impact of the input-tailored missing-aware prompt's length on model performance. The illustration reveals a discernible U-shaped trend where an increase in prompt length enhances model performance. However, beyond a certain threshold, excessively long prompts tend to blur learned features, resulting in performance degradation. The peak performance is observed when the prompt length is set at 20, implying that the prompt length should also be considered a compromise factor.

### 5.5 Limitations and Future Works

In the future, we plan to improve the model's adaptability to handle signals of varying lengths while expanding its capacity to integrate additional view signals from CVS. Additionally, we hypothesize that integrating frequency domain and demographic information into the model could improve the downstream tasks' performance.

Beyond the prompt techniques introduced in this article, constructing shared and specific features to handle view missing represents a promising technical path [63]. As our pretraining tasks involve cross-view signal reconstruction, this pushes us to explore the feasibility of incorporating this technique further. It is important to note that while the term 'multi-view' is used here, 'multi-view' and 'multimodal' are frequently used interchangeably in physiological signal processing. Therefore, our proposed method should not be confined solely to the domain of multi-view.

## 6. Conclusion

In this study, we observe a prevalent tendency in existing methodologies to interpret multi-view signals as mere multi-channel signals, leading to the issue of view confusion and impeding the extraction of crucial representations from multi-view data. We propose the VCT, a novel approach founded on view-centric tokenization and classical MSA layers. This framework enables a more effective capture of temporal and cross-view interactions inherent in multi-view signals. Moreover, our proposed M2AE leverages unlabeled multi-view signal pairs, empowering VCT to acquire comprehensive representations and significantly enhancing downstream task performance. This paradigm demonstrates the potential of utilizing a broader range of supervised sources, optimizing the efficiency of unlabeled data utilization, and reducing reliance on costly medical labeling. To tackle an overlooked yet significant missing-view problem, we design input-tailored and attention-tailored missing-aware prompt techniques to denote different missing-view scenarios within pretrained models. This method minimizes the need for extensive model finetuning, significantly reducing reliance on computational resources. Such an approach is paramount in the current burgeoning data volumes and escalating model complexities. We hope our strong results and the simple methods can facilitate further research in MVF and missing-view issues.